\documentclass[lettersize,journal]{IEEEtran}
\usepackage{amsmath,amsfonts}
\usepackage{algorithmic}
\usepackage{algorithm}
\usepackage{array}
\usepackage[caption=false,font=normalsize,labelfont=sf,textfont=sf]{subfig}
\usepackage{textcomp}
\usepackage{stfloats}
\usepackage{url}
\usepackage{verbatim}
\usepackage{graphicx}
\usepackage{cite}
\usepackage{multirow}

\begin{document}

\title{TrajAware: Graph Cross-Attention and Trajectory-Aware for Generalisable VANETs under Partial Observations}

\author{Xiaolu Fu, Ziyuan Bao, Eiman Kanjo
\thanks{This work will be submitted to the IEEE for possible publication.
Copyright may be transferred without notice, after which this version may
no longer be accessible.

Xiaolu Fu is an AI research engineer at Unicom Data Intelligence, China Unicom, Hangzhou, China (fuxl67@chinaunicom.cn), and a former student of the Computing Department, Imperial College London, London, UK (email: andy.fu23@alumni.imperial.ac.uk).

Ziyuan Bao is an independent researcher and a former MSc student of the Computing Department, Imperial College London, London, UK (email: ziyuan.bao23@alumni.imperial.ac.uk).

Eiman Kanjo is a Professor with Pervasive Sensing \& TinyML and the Head
of the Smart Sensing Lab at Nottingham Trent University, Nottingham, UK
(email: eiman.kanjo@ntu.ac.uk); and Provost’s Visiting Professor in tinyML
at Imperial College London, London, UK (email: e.kanjo@imperial.ac.uk).
}}



\maketitle

\begin{abstract}

Vehicular ad hoc networks (VANETs) are a crucial component of intelligent transportation systems; however, routing remains challenging due to dynamic topologies, incomplete observations, and the limited resources of edge devices. Existing reinforcement learning (RL) approaches often assume fixed graph structures and require retraining when network conditions change, making them unsuitable for deployment on constrained hardware. We present \textbf{TrajAware}, an RL-based framework designed for \textbf{edge AI deployment in VANETs}. TrajAware integrates three components: \textbf{(i) action space pruning}, which reduces redundant neighbour options while preserving two-hop reachability, alleviating the curse of dimensionality; \textbf{(ii) graph cross-attention}, which maps pruned neighbours to the global graph context, producing features that generalise across diverse network sizes; and \textbf{(iii) trajectory-aware prediction}, which uses historical routes and junction information to estimate real-time positions under partial observations. We evaluate TrajAware in the open-source SUMO simulator using real-world city maps with a leave-one-city-out setup. Results show that TrajAware achieves near-shortest paths and high delivery ratios while maintaining efficiency suitable for constrained edge devices, outperforming state-of-the-art baselines in both full and partial observation scenarios. 
\end{abstract}


\section{Introduction}

\begin{figure}
    \centering
    \includegraphics[width=1\linewidth]{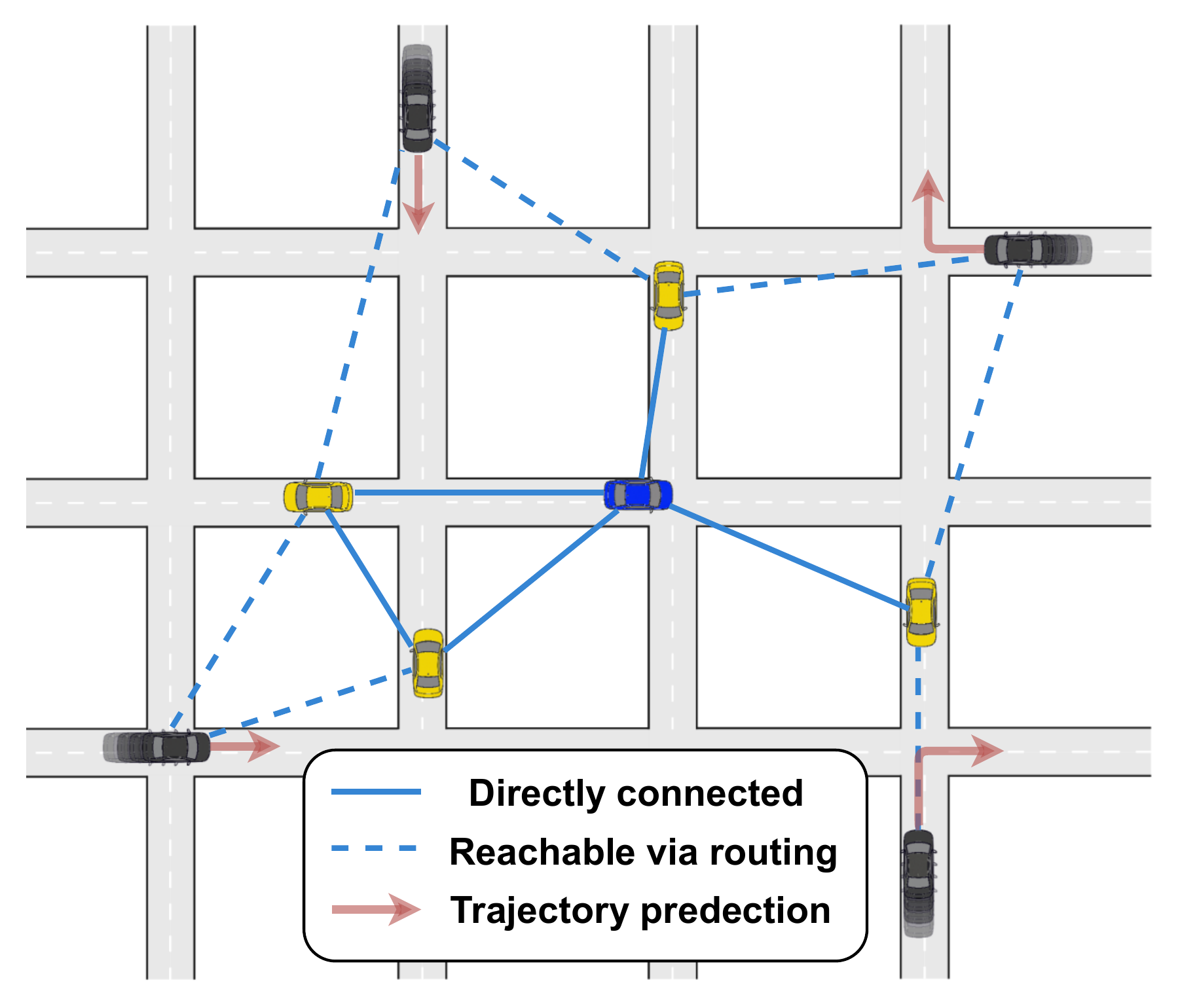}
    \caption{Conceptual street-view diagram of a VANET. The blue vehicle represents the ego vehicle, yellow vehicles are within its communication range, and the remaining vehicles are unobservable while in motion.}
    \label{fig:vanetframe}
\end{figure}

\IEEEPARstart{C}{ommunication} and routing are challenging in a vehicular ad hoc network (VANET) \cite{5GvsDSRC}, as vehicles can observe only part of the network, and the network's structure shifts rapidly; a previously obtained observation may soon become obsolete (as shown by \textbf{\textbf{Figure~\ref{fig:vanetframe}}}). Although compared to classical software algorithms, RL routing algorithms can potentially deal with more complex objectives (e.g., optimising delay while minimising the bandwidth overhead) \cite{FANET_HELLO}, the problems of partial observation and network dynamics put a strain on the RL routing models.

Several studies have shown that graph neural networks (GNNs) generalise better on routing tasks compared to other neural networks like multilayer perceptrons (MLPs) \cite{Baseline_DRL_vs_GRL, DRL_vsGRL, GDDR_action_space_prunning, Recurrent_GRL, wc_gnn_review}. Because GNNs process the topological information of the network, they do not have to overfit to a certain set of network structures. Theoretically, a GNN-based routing algorithm could be generalised to highly dynamic VANET environments and unseen network graphs. However, there is still a significant gap between experimental assumptions and the complexities of real-world VANET environments: existing research on GNN-based routing often tests their methods on graphs with a fixed small number of nodes and d-regularity (usually a small d); whereas in real-world environments, neither the node number nor the node degree is fixed, and they could potentially be significant. Moreover, GNN-based routing research rarely address partial observations, making them less compatible with VANET routing.

GNNs can handle inputs of arbitrary sizes, but the outputs' data structure remains graphs. To bridge the gap between GNNs' outputs and the action space of a routing policy, a common approach is to concatenate all node features into one vector, then several MLP layers reduce the vector's dimension to the number of next-hop choices. Since MLPs require fixed input and output sizes, a routing model that generalises across different network sizes and node degrees must initialise the MLP with large dimensions and pad inputs or outputs when the actual sizes are smaller. However, this approach introduces both \textbf{the curse of dimensionality} and \textbf{the long tail problem}.


We propose a novel system that addresses the above problems. Firstly, agents proactively broadcast and exchange their knowledge with neighbours to form more complete but outdated observations, and a trajectory prediction algorithm estimates the real-time positions of other vehicles. We design an action space pruning approach, which alleviates the long tail problem and the curse of dimensionality by ignoring neighbours that are less likely to contribute to the routing task.
We adopted the cross-attention from the transformer \cite{attentionIsAllUNeed} to address the curse of dimensionality, as well as enhanced the model's ability to learn inductively. Our main contributions are listed in the following:
\begin{itemize}
    \item Proposed TrajAware, a novel reinforcement learning algorithm that generalises across graphs with varying node degrees and network sizes, addressing the limitations of prior methods constrained to fixed topologies.

    \item Enriched VANET optimisation with Graph Cross-Attention, improving scalability and inductive learning compared to conventional graph neural network–based approaches.

    \item Tackled the challenge of partial and outdated observations in VANETs by integrating trajectory-aware prediction, enabling vehicles to estimate real-time positions of neighbours and maintain robust decision-making.

    \item Extended the evaluation of existing methods to more complex and realistic environments, demonstrating generalisability beyond simplified benchmarks. We validated TrajAware with the open-source SUMO traffic simulator and real-world city road networks.
\end{itemize}
To the best of our knowledge, although there exists research that utilise cross-attention on GNNs \cite{nature_gca}, we are the first to apply a similar network architecture on routing for VANET and demonstrate its superior generalisability.

The remainder of this paper is organised as follows. Section II provides an overview of related work in VANET routing, including graph-based reinforcement learning approaches, attention mechanisms, and trajectory prediction. Section III details our proposed TrajAware methodology, describing how we integrate action-space pruning, cross-attention, and trajectory forecasting into a GraphSAGE-based routing framework. Section IV presents the evaluation of TrajAware in realistic traffic simulations, demonstrating the generalisation of the model to unseen networks and comparing its performance against other methods under complete and partial observations. Finally, Section V concludes the paper with a summary of findings and discusses a potential direction for future research.


\section{RELATED WORK}
\subsection{Graph Neural Networks}
Inspired by the success of convolutional neural networks (CNNs) in many domains, computationally efficient convolution on graph structures has been proposed \cite{GCN_efficient}. Like CNNs, graph convolutional neural networks (GCN) aggregate neighbouring nodes' features. Unlike CNNs, classical GCNs do not have a learnable kernel to weigh neighbours' features; a learnable dense matrix ($W$) projects each convolution result and then adds a bias ($b$).

Let $x_i^k$ be the $\text{i}^{\text{th}}$ node feature of the $\text{k-1}^{\text{th}}$ layer or the input to the $\text{k}^{\text{th}}$ layer. A graph convolution \cite{gcn_define} is formally defined as:
\begin{equation}\label{eq:GCN_1}
    x_i^k = W^T (\sum_{j} x_j^{k-1} \cdot \frac{A_{ij}}{\sqrt{deg(x_i) \times deg(x_j)}}) + b
\end{equation}
$W$ is a weight matrix, $b$ is the bias, $A$ is the adjacency matrix, and $A_{ij}=1$ implies $x_j$ is a neighbour of $x_i$. The sum is normalised by node degrees, $deg(\cdot)$, which is the number of the node's neighbours.

To fully utilise GPU parallel computing, let $X^{k-1}$ be the input node features to the $\text{k}^{\text{th}}$ layer. The output $X^{k}$ can be computed with:
\begin{equation}
    X^{k} = \hat{A}X^{k-1}W^k + b^k
\end{equation}
Where $\hat{A}$ is a renormalised adjacency matrix, and it is calculated with:
\begin{equation}
    \begin{gathered}
    \hat{A} = \Tilde{D}^{-\frac{1}{2}}\Tilde{A}\Tilde{D}^{-\frac{1}{2}} \\
    \Tilde{A} = A + I \\
    \Tilde{D}_{ii} = \sum_j \Tilde{A}_{ij}
    \end{gathered}
\end{equation}
The original paper \cite{gcn_define} introduced renormalisation to stabilise gradients. $\tilde{A}$ is the adjacency matrix with an additional self-connection. 
$\tilde{D}$ is a matrix for node degrees.

When multiple GCN layers are stacked, node features are aggregated from increasingly distant neighbours. This often leads to the over-smoothing problem, where node representations become indistinguishable from one another. Several approaches have been proposed to mitigate this issue. For example, the Jumping Knowledge network \cite{JK} and GraphSAGE \cite{graphsage} preserve intermediate representations from each layer and combine them through concatenation, pooling, or an additional neural network.

Like other models, GCNs could be applied in RL \cite{GCRL_review}. In the original deep Q-network (DQN) research, the agent plays Atari games and takes images as input \cite{DRL2013}; CNN naturally excels in image processing and is a part of their model. Similarly, GCN could be an essential part of the algorithm when the inputs are graphs. Since communication networks can be represented with graphs, this project involves graph reinforcement learning (GRL). 

\subsection{Reinforcement Learning}
Reinforcement learning (RL) is particularly suitable in scenarios where ground truth is difficult to obtain or even define. Instead of relying on explicit labels, RL employs a carefully designed reward function as feedback, and agents attempt to collect the most rewards. The routing problem is an application of RL since the consequences of actions cannot be known immediately. Still, rewards can be designed to reflect performance metrics such as delay and throughput.

RL methods aim to find a policy $\pi$ that maximises total discounted rewards \cite{RLHistory}. A value function of a state under the policy $V^\pi$ is defined as:
\begin{align}\label{eq:V_value}
    V^\pi(s) 
    &= \mathbb{E}\!\left[\sum_{k=0}^\infty \gamma^k r_{t+k+1}\,\middle|\,s_t = s\right] \nonumber \\
    &= \sum_{a \in A} \pi(s, a) 
       \sum_{s' \in S} P^a_{ss'} \big( R^a_{ss'} + \gamma V^\pi(s') \big)
\end{align}
Where $\gamma$ is the discount factor, indicating how important is a future reward over an immediate reward, a small $\gamma$ makes the agent short-sighted; $r_t$ is the reward received at time $t$ and $s_t$ is the state; $\pi(s, a)$ is the probability of selecting action $a$ in state $s$ under policy $\pi$; $s'$ is the resulting state of an action, $P^a_{ss'}$ and $R^a_{ss'}$ denotes the probability and reward of reaching state $s'$ after taking action $a$ at state $s$.

Alternatively, one could assign value to state-action pairs, the value for selecting action $a$ at state $s$, namely $Q^\pi$:

\begin{align}\label{eq:Q_value}
    Q^\pi(s, a) 
    &= \mathbb{E}\!\left[\sum_{k=0}^\infty \gamma^k r_{t+k+1}\,\middle|\,s_t = s, a_t=a\right] \nonumber \\
    &= \sum_{s' \in S} P^a_{ss'} \big( R^a_{ss'} + \gamma V^\pi(s') \big)
\end{align}
A desirable agent should approximate true state values with $V^\pi$ (or approximate true Q-values with $Q^\pi$), and optimising the decisions should be equivalent to maximising the values. In routing tasks, a typical state $s$ would be the observed network topology, and an action $a$ selects the next hop.

In the real world, the state space is often ample or infinite, and storing all values is infeasible. To solve this problem, DRL methods with neural networks have been proposed \cite{DRL2013, DRL2015}. Since non-linear neural networks are universal approximators \cite{universal_approximators}, they could approximate the value function. This concept has also been adopted in traffic-related problems \cite{RL_AutoDrive_survey}.

\subsection{GRL Routing}
Several recent studies have adopted GRL on the routing task. Bhavanasi et al. \cite{Baseline_DRL_vs_GRL} compared a traditional DRL method with their GRL method in terms of generalisability. They concluded that GRL generalises better on graphs with various topologies, but their GRL model still requires retraining under certain conditions. In their experimental environment, on average, each node has only two neighbours. The argument that GRL outperforms non-graph DRL in routing generalisability has been further supported by Almasan et al. \cite{DRL_vsGRL}. In contrast, Casas-Velasco et al. claimed that their non-graph model required more than 10 seconds to recover from a topology change \cite{DRL_QoS}, which is already an improvement compared to an early study. All the above studies assume the graph is provided by default. However, a router cannot observe the whole graph in many routing tasks. Weil et al. addressed partial observation by integrating reactive recurrent message passing with GRL \cite{Recurrent_GRL}. However, their model has only been trained and tested on networks with 20 nodes, and each node always has a degree of 3, resulting in a limited range of unique topologies (according to the House of Graphs \cite{house_of_graphs}). Thereby, each graph topology in their test set is very likely to be seen during the training.

\subsection{Cross-Attention with GNNs}
The cross-attention mechanism was initially designed as a component of the Transformer \cite{attentionIsAllUNeed}. The attention mechanism is particularly effective at modelling long-range dependencies in sequences, which is well-suited for our needs of processing a list of node features. While attention's permutation invariance and equivariance are undesired in natural language tasks due to the importance of word order, this limitation is not present in our use case, where nodes are inherently unordered. Shen et al. \cite{nature_gca} proposed a novel architecture combining cross-attention with GNNs to integrate node features and topology information, they also provided a concise review of related work in this direction. As they pointed out, most existing approaches employ cross-attention to bridge graph structures with non-graph structures.

\subsection{Trajectory Prediction}
We aim to predict the vehicle's future position given historical data. Trajectory prediction has been intensively studied. Since the historical trajectory only provides information of an object's short-term momentum and the style of movement, extra input is often required. For example, Xu et al. \cite{imputation_prediction} introduced a method that combines a graph neural network with a variational RNN to forecast the positions of all objects simultaneously.  The GNN captures interactions between objects, which is particularly relevant in their experiments on sports games, where the movement of each object is highly interdependent. Kim et al. \cite{Long-Term_trajectory} pointed out that errors tend to build up in multi-step prediction. They proposed a ``link projection" method to alleviate this problem, in which an algorithm projects the prediction onto the nearest road whenever the model predicts a position. Additionally, they proposed a binary encoding for the vehicle's surrounding situations, including traffic lights, bus stops, intersections, and speed limits, as these situations typically influence drivers' decisions.

The method of Kim et al. \cite{Long-Term_trajectory} has profoundly inspired us, and we adopt both the link projection method and the binary encoding of road context in our framework. In addition, recognising that a vehicle’s future trajectory is often guided by its intended route or destination, we incorporate planned paths as input to our model. We assume such information can typically be obtained from navigation systems or autonomous driving agents.

\begin{figure*}[!t]
\centering
\includegraphics[width=7in]{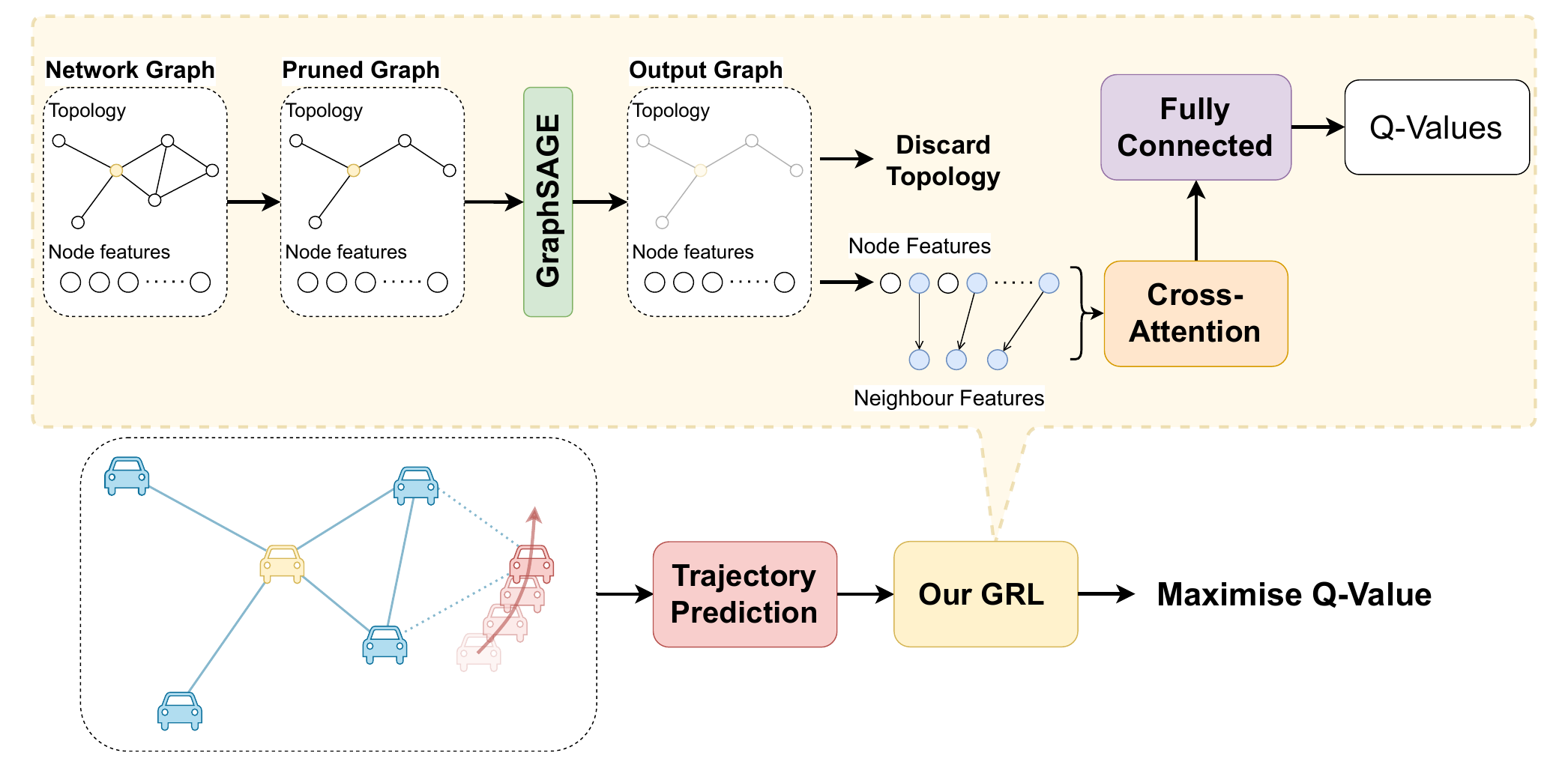}
\caption{Our graph reinforcement-learning method, TrajAware. A graph is represented by a tensor that encodes the topology (adjacency matrix or edge index) and a tensor of node features. Our final layer is a fully connected layer instead of an MLP, as it does not have a non-linear layer afterwards.}
\label{fig:grl}
\end{figure*}

\section{Methodology}
We present TrajAware, a routing method designed to generalise across diverse network environments without retraining. Our GRL model combines action space pruning and attention mechanisms, built upon a GraphSAGE \cite{graphsage} backbone. Our GRL model applies not only to static networks with complete observations but also to dynamic VANET routing tasks when integrated with our trajectory prediction system. \textbf{Figure \ref{fig:grl}} demonstrates the pipeline of the TrajAware.

Our GRL process begins with an action space pruning algorithm, which significantly reduces the set of candidate actions. The resulting pruned graph is passed to the GraphSAGE module, which aggregates node features and generates a graph representation. Next, we extract the neighbouring nodes (not including the pruned neighbours) of the current sender node to form a subset of node features. Both the full node set and the neighbour subset are input into a cross-attention layer, which produces a list of vectors, each refers to one neighbour. These vectors are then concatenated and passed through a final fully connected (FC) layer to produce the Q-values. Since the attention mechanism operates solely on node features, the topology information becomes unnecessary at this stage and is discarded before the cross-attention layer is applied.

Given that existing communication standards, such as DSRC, require vehicles to periodically broadcast their directions and positions \cite{DSRC_USA}, we assume it is feasible to transmit a reasonable amount of additional information, including the vehicle’s planned path and local observations.

Although vehicles share their observations, information from distant vehicles often arrives with a delay and may become outdated. Delayed data is stored as historical trajectories. These are used to predict vehicles’ real-time positions. Given the sequential nature of vehicle trajectories, we employ a Gated Recurrent Unit (GRU) \cite{GRU} for this task. While trajectory prediction plays a vital role in our system, it is not the primary focus of this study. Instead, we emphasise model deployability and therefore adopt the parameter-efficient GRU as a lightweight yet effective solution.

The remainder of this section introduces the three key components of our approach: action space pruning, cross-attention, and trajectory prediction.

\subsection{Action Space Pruning}
The routing decision resembles a classification problem: given the current state, a router must select one neighbour (class) as the next hop from $n$ neighbours (classes). Real-world networks, especially VANETs, exhibit huge variance in node degrees. Some routers (nodes) naturally have more neighbours and a larger node degree. We need to initialise the neural network with the maximum possible number of neighbours, denoted as $n_{max}$. For routers with fewer than $n_{max}$ neighbours, we pad the input to match the expected dimensionality. However, this leads to an imbalance during training. Specifically, when a router has only $m < n_{max}$ neighbours, the padded entries (i.e., neighbour indices $i$ such that $m<i\leq n_{max}$) are never selected as the correct class. Consequently, in the training data, the $(i+1)^{th}$ class consistently appears no more frequently than the $i^{th}$ class, resulting in a highly imbalanced classification, also known as the long tail problem, which hampers learning.

Moreover, $n_{max}$ is often large, leading to the curse of dimensionality. Although the routing problem differs from the classification problem, a routing agent predicts Q-values rather than probabilities, and it is trained using reinforcement learning rather than supervised learning. The problems of biased distribution and the curse of dimensionality also apply to routing tasks. However, rarely do routers have $n_{max}$ neighbours, nor are all neighbours equally relevant for routing decisions. We found that reducing the action space and balancing the action choices are essential.

\begin{figure}[!t]
\centering
\includegraphics[width=3.5in]{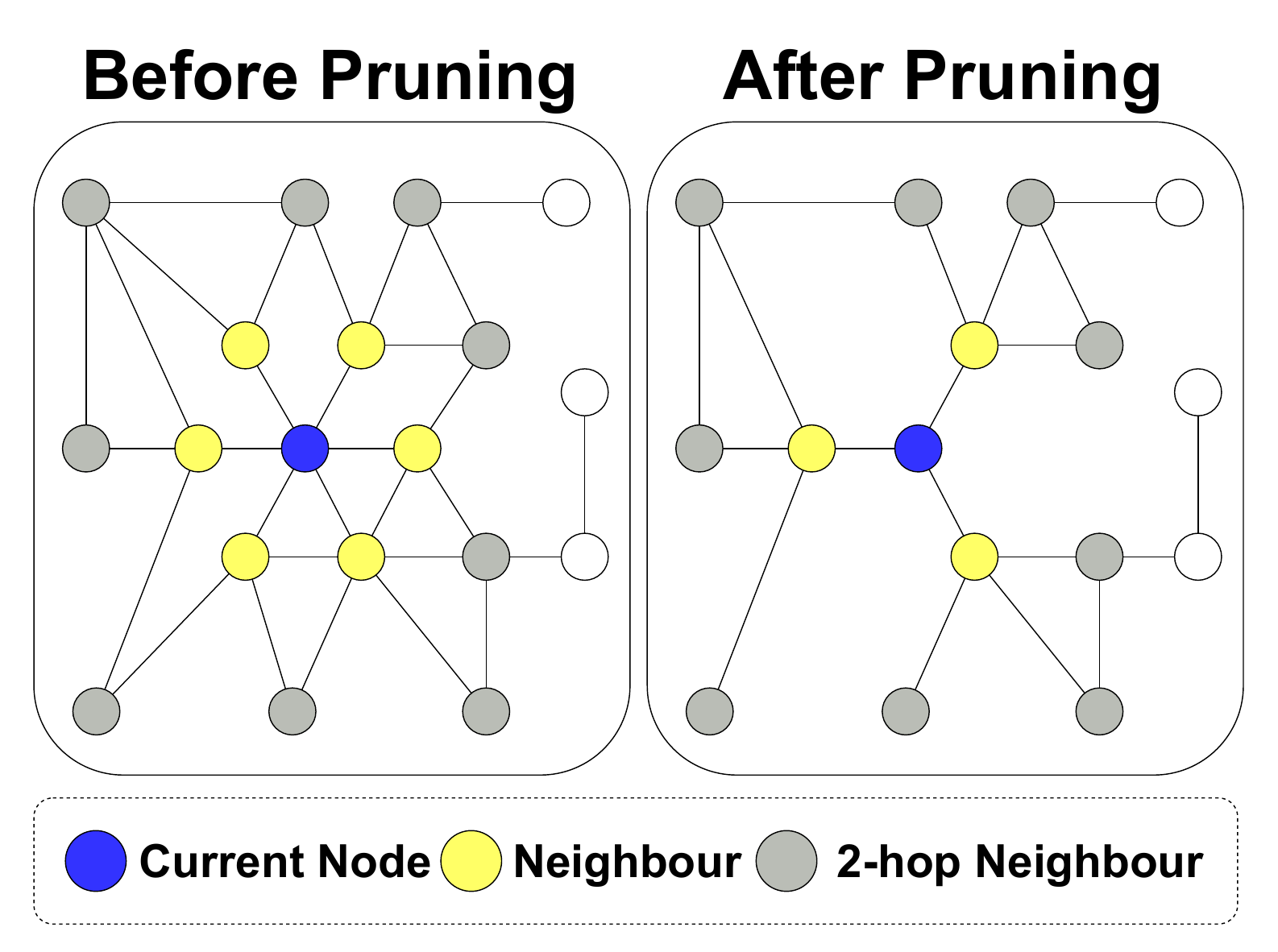}
\caption{An illustration of action space pruning. The left side shows the current node and all reachable nodes. The right side demonstrates the effect of pruning, where only the most representative neighbours that can reach all 2-hop neighbours are retained.}
\label{fig:node_pruning}
\end{figure}

Some neighbours may not contribute to routing and could be ignored. This is because:
\begin{itemize}
    \item If a neighbour is the destination of the routing, the agent can directly relay the message to it. In this case, no sophisticated routing algorithm is needed.
    \item Otherwise, the destination is at least two hops away from the current agent. In this case, the agent only needs to consider a set of direct neighbours that can reach all 2-hop neighbours.
\end{itemize}
While this approach may intuitively raise concerns about potential bottlenecks or reduced bandwidth, our experimental results suggest otherwise. Specifically, we find that the positive impact of pruning, which lies in eliminating suboptimal next-hop choices and simplifying the learning problem, outweighs its potential negative effects. Moreover, bandwidth occupancy in real-world networks is highly dynamic and only partially observable. As a result, even if an agent is given more routing options, it often cannot reliably determine which choice would lead to better bandwidth performance. \textbf{Figure \ref{fig:node_pruning}} shows an example of node pruning. After removing three neighbouring nodes, the current node can still reach all other nodes with the same number of hops.

It is computationally challenging to select the smallest set of neighbours since there are $2^n$ possible ways of selection, where $n$ is the number of neighbours. We do not aim to select the optimal set. Instead, we rank neighbours by the number of links to 2-hop neighbours, greedily check neighbours in this order, and remove those that do not provide new paths to 2-hop neighbours. This can already closely approximate the optimal pruning, and as a result, we reduced the action space from 21 to 8 in our experimental environment. In rare cases where a router still has more than eight neighbours after pruning during testing, we randomly select eight and discard the rest. Given the infrequency of this scenario, its impact on performance is negligible.

Notably, the node degree distribution remains highly skewed after pruning, with only a small fraction of vehicles retaining eight neighbours. To address this imbalance, we slightly modified the pruning strategy:
\begin{itemize}
    \item No pruning is performed if a vehicle has no more than eight neighbours.
    \item If a vehicle has more than eight neighbours, we remain eight neighbours, even if some of them do not provide new paths to 2-hop neighbours.
\end{itemize}

\begin{figure}[!t]
\centering
\includegraphics[width=3.5in]{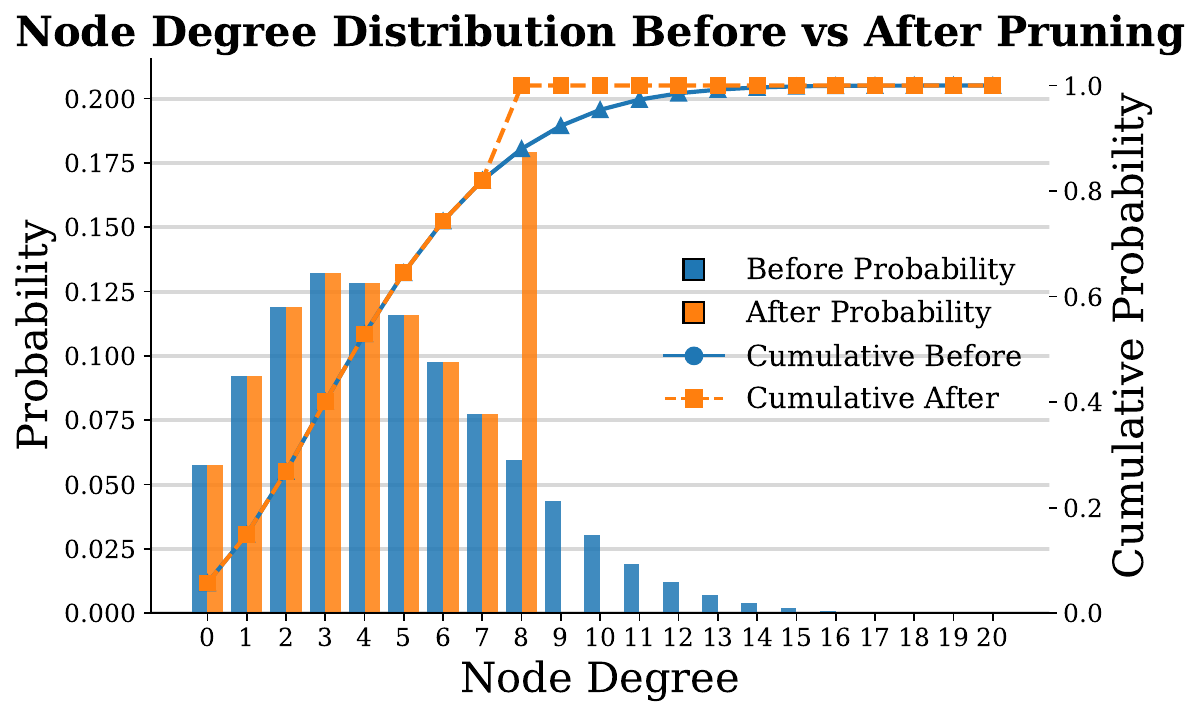}
\caption{Node degree distributions before and after pruning. The modified greedy strategy limits the maximum degree to 8 and balances the distribution, improving training convergence and generalisability.}
\label{fig:before_vs_after_pruning}
\end{figure}

As a result, a more balanced distribution of node degrees is obtained, as shown in \textbf{Figure~\ref{fig:before_vs_after_pruning}}. This node pruning helped with training convergence by balancing the sample distribution. This method has also improved the model's generalisability on graphs with large node degrees.

\subsection{Cross Attention}
We identified a significant challenge in training the routing agent, compared to some other classification tasks or agent problems. In classification, each class typically corresponds to a well-defined and meaningful data distribution, such as images of a flower or textual descriptions of a movie. Similarly, in many standard reinforcement learning settings, such as Atari games, each action often carries a consistent, interpretable meaning; for example, a specific action might always correspond to ``move forward" in a video-game-based environment.

In contrast, this consistency no longer applies in routing. Each action corresponds to selecting a neighbour node, but the identity and order vary across different situations. For instance, ``action 1" might refer to completely different neighbours depending on the current agent’s location. Furthermore, there is no inherent or meaningful ordering of neighbours, which makes it difficult to assign stable semantics to individual actions. As a result, the model must learn not only which action to take, but also how to interpret what each action means within the local context.

Our design also draws inspiration from the Transformer architecture, particularly its ability to handle variable-length input. Moreover, the cross-attention mechanism is permutation invariant and equivariant, which aligns well with the nature of routing tasks, as we should ignore node order.

A cross-attention layer takes two lists of node features as input $S$ and $S'$, where $S$ represent a set of neighbours and $S$ is extracted from the set of all nodes $S'$.
\begin{equation}
    \begin{gathered}
        S = [s_1, s_2, s_3, \dots, s_n] \\
        S' = [s_1', s_2', s_3',\dots, s_m']
    \end{gathered}
\end{equation}
The following explanation of the attention mechanism strictly follows the definition of the original paper \cite{attentionIsAllUNeed}. Firstly, a learnable matrix $W_q$ projects each vector in $S$ to form a query $q_i$, another two matrices $W_k$ and $W_v$ project vectors in $S'$ to form key $k_i$ and value $v_i$. By stacking queries, keys and values vectors respectively, we get :
\begin{equation}
    \begin{gathered}
        Q = [q_1, q_2, q_3, \dots, q_n]^T \quad\qquad \text{where } q_i = W_q \cdot s_i \\
        K = [k_1, k_2, k_3, \dots, k_m]^T \quad\qquad \text{where } k_i = W_k \cdot s_i' \\
        V = [v_1, v_2, v_3, \dots, v_m]^T \quad\qquad \text{where } v_i = W_v \cdot s_i'
    \end{gathered}
\end{equation}
An attention score is the dot product similarity between a query and a key. The definition of the attention score matrix $A$ is:
\begin{equation}
    A = \sigma(\frac{Q\cdot K^T}{\sqrt{d_h}})
\end{equation}
where $\sigma$ is the softmax activation function. Each query and key vector has the same hidden dimension $d_h$, and this hidden dimensionality is used for normalisation.

The attention module's final output matrix $O$ is the product of A and V:
\begin{equation}\label{eq:attention}
    \text{Attention($S$, $S'$)} = O = A\cdot V
\end{equation}
If we shuffle the input $S$ to the attention module, the output will be shuffled in the same order, so attention is permutation equivariant. If we shuffle the input $S'$, the output will not be affected; attention is also permutation invariant:
\begin{equation}
    \begin{gathered}
        \text{Attention}(f(S), S') = f(\text{Attention}(S, S')) \\
        \text{Attention}(S, f(S')) = \text{Attention}(S, S')
    \end{gathered}
\end{equation}
The cross-attention layer outputs a list of hidden node features with an equivalent length to $S$. The attention mechanism naturally learns which neighbour in $S$ attends to which node in the context set $S'$. Moreover, the \text{$i^{th}$} output will always correspond to the \text{$i^{th}$} neighbour. As a result, the cross-attention layer can inductively generalise to unseen graphs, since it operates over sets without relying on a fixed graph structure or ordering.

The output of the cross-attention layer is an $n$ by $h$ tensor. Where $n$ is the number of neighbours, each neighbour has a hidden vector of size $h$. This tensor is then flattened and passed to an FC layer, which produces an output vector of size 
$n$. Since the attention mechanism already captures pairwise interactions among neighbours, the FC layer primarily serves as a dimensionality reduction and mapping to Q-values. As a result, the order of the input neighbours has minimal effect on the output.

\subsection{Trajectory Prediction Algorithm}
In a partially observable VANET routing environment, we employ a GRU model to predict the real-time positions of vehicles. We assume that each agent is aware of its planned path, broadcasts its local observations, and integrates them with observations received from neighbouring vehicles.

The vehicle's direction uncertainty complicates vehicular trajectory prediction. The model needs to know the vehicle's intended path; without this knowledge, the model must infer whether the driver will turn or continue straight at intersections. Additionally, predicting a vehicle's trajectory on a curved road is challenging unless the model overfits a specific map.

To address this, we designed a segment node mechanism. A segment node marks the end of a straight road segment, typically at a traffic junction. In map data, curved roads are usually approximated by a series of line segments, each ending at a segment node. 

Two consecutive segment nodes are enough to define a direction change. The input to the GRU model consists of a 6-dimensional vector, which includes the x and y coordinates of the vehicle's current position, as well as the coordinates of the two subsequent segment nodes. We observed that segment nodes are unevenly distributed on the map, with some roads being short and curved while others are long and straight. This uneven distribution causes unstable training, so we also split long, straight roads with segment nodes.

To reduce the bandwidth overhead, a car could broadcast only the junctions it will visit, and the segment nodes on curved roads could be computed from a locally stored map.

We are inspired by Kim et al. \cite{Long-Term_trajectory}, who project predicted trajectories onto the nearest road to reduce error. As a supplementary point, this projection is differentiable and parallelisable, enabling direct integration into a batch learning pipeline.
\section{EVALUATION}
To simulate the real VANET routing environment, we used the Simulation of Urban Mobility (SUMO) \cite{SUMO}. SUMO offers tools to create environments from scratch or load real traffic networks from OpenStreetMap \cite{OpenStreetMap}. To ensure data diversity, we selected six cities from different continents: Chiang Mai (Thailand), Christchurch (New Zealand), Edinburgh (UK), Nairobi (Kenya), Portland (US), and Rio de Janeiro (Brazil). These cities were chosen to represent a balanced mix of developing and developed regions.

To evaluate the generalisability of our method, we adopted a leave-one-city-out approach: models were trained using data from five of the selected cities and tested on the sixth. In our main experiment, we used Edinburgh as the held-out test environment. This setting forces the models to perform in a completely unseen, real-world-like scenario without retraining, serving as a rigorous test of their generalisation across urban traffic patterns.

\begin{figure}[!t]
\centering
\includegraphics[width=3.2in]{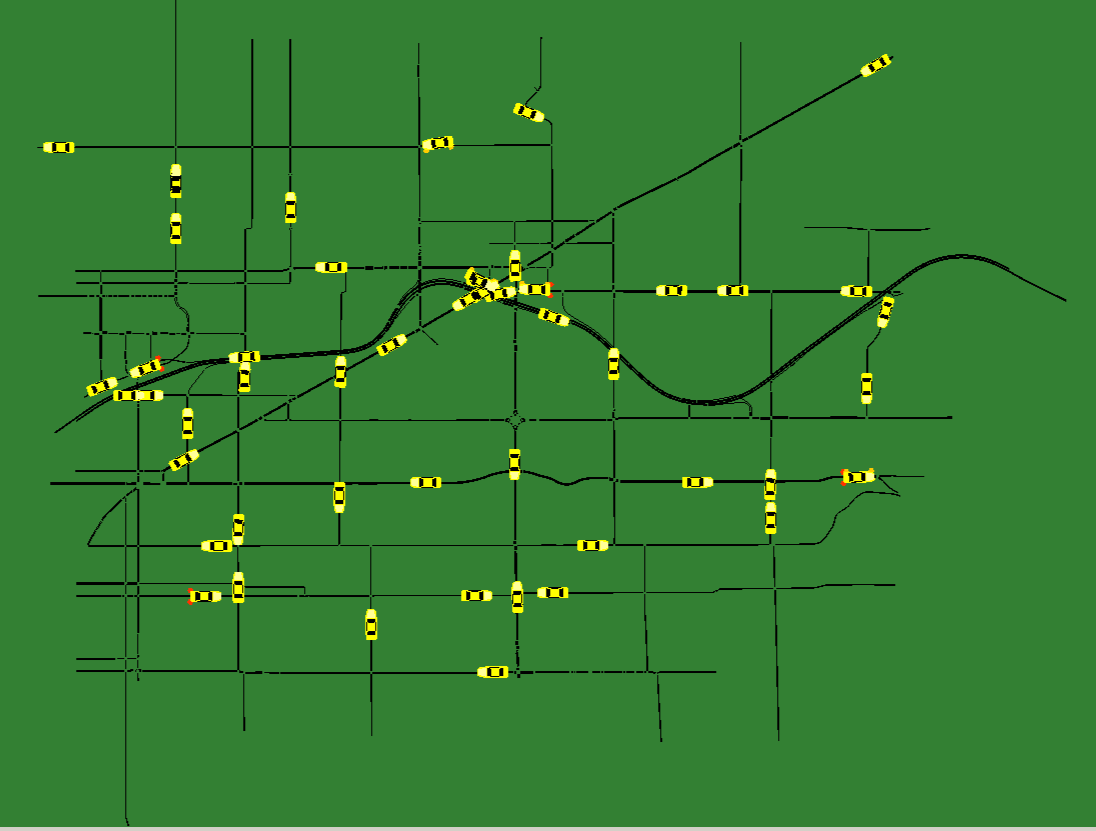}
\caption{Road network of Portland in the simulation environment. The black lines represent roads, and the yellow ones represent vehicles.}
\label{fig:Screenshot}
\end{figure}

\textbf{Figure~\ref{fig:Screenshot}} shows the simulated environment of Portland. Each map covers an area of approximately 7000 $\times$ 6500 meters, though the exact dimensions vary slightly depending on how the road network is cropped. To maintain computational efficiency, only major roads are included in the simulation, as modelling all street-level details across a large area would be expensive. Additionally, we manually added a few key roads to approximate some excluded streets.

We used SUMO's built-in trip generator to create simulated vehicles. This tool takes two key parameters: the generation duration and the density factor. The generation duration defines the time period (in real-world seconds) over which trips are generated from the start of the simulation. The number of vehicles produced is proportional to both the density factor and the underlying road network density.

In our experiments, the generation duration was set to 10,000 seconds for all cities. The density factor was adjusted per city to account for variations in road density. To ensure vehicles could establish connections while avoiding excessive computational overhead, we controlled the number of simultaneously active vehicles to remain between approximately 40 and 70 in each step.

Although the connection range of any wireless method is highly dependent on the environment. We assume a signal range of 800 meters, slightly below the theoretical upper limit of DSRC \cite{DSRC_USA}. If two vehicles are within 800 meters of each other, we assume they are connected. Please note that the SUMO simulator has a time resolution of one second. In the following experiments, one step refers to one second.

\subsection{Complete Observation}
We first evaluated GRL models with complete observations. We extracted vehicle positions directly from the SUMO simulation and computed the network structure every five steps (because consecutive steps tend to yield the same network structure). In the congestion mode, the bandwidth configuration follows Weil et al. \cite{Recurrent_GRL}, where each link has a capacity of 1 and each packet has a random size between 0 and 1.

\begin{table}[!t]
\renewcommand{\arraystretch}{1.3}
\centering
\caption{Ablation results under congestion and no congestion. \checkmark indicates the component is used.}
\label{tab:complete}
\begin{tabular}{l|cc|ccc}
\hline
\multirow{2}{*}{\textbf{Mode}} & \multicolumn{2}{c|}{\textbf{Components}} & \multicolumn{3}{c}{\textbf{Metrics}} \\ \cline{2-3}\cline{4-6}
 & \textbf{Pruning} & \textbf{Attention} & \textbf{Avg. SPR} & \textbf{Avg. PSPR} & \textbf{RR} \\ \hline
\multirow{4}{*}{\rotatebox{90}{\scriptsize No Congestion}} 
 & \checkmark & \checkmark & \textbf{1.0031} & \textbf{1.0565} & \textbf{0.9722} \\
 & $\times$ & \checkmark & 1.0033 & 1.0875 & 0.9630 \\
 & \checkmark & $\times$ & 1.4898 & 4.4799 & 0.1968 \\
 & $\times$ & $\times$ & 1.1080 & 4.5713 & 0.1672 \\ \hline
\multirow{4}{*}{\rotatebox{90}{\scriptsize Congestion}} 
 & \checkmark & \checkmark & 1.1272 & \textbf{1.1910} & \textbf{0.9652} \\
 & $\times$ & \checkmark & \textbf{1.1244} & 1.2014 & 0.9525 \\
 & \checkmark & $\times$ & 1.6732 & 4.6765 & 0.1637 \\
 & $\times$ & $\times$ & 1.5192 & 4.7588 & 0.1493 \\ \hline
\end{tabular}
\end{table}

We evaluated three metrics: shortest path ratio (SPR), penalised shortest path ratio (PSPR), and reached ratio (RR):
\begin{equation}
\begin{gathered}
    \text{SPR} = \text{no. hops used} / \text{shortest path} \\
    \text{PSPR} = 
        \begin{cases}
            \text{SPR} & \text{Packet delivered} \\
            \text{TTL}/\text{shortest path} & \text{Packet dropped} \\
        \end{cases} \\
    \text{RR} = \text{no. packets reached} / \text{no. packets sent}
\end{gathered}
\end{equation}

The average SPR is closely related to both delay and throughput. Since these metrics also depend on the shortest path in a routing task, SPR directly reflects the probability that the model makes a suboptimal decision at each step. We also monitor the reachability ratio (RR), defined as the percentage of packets that successfully reach their destination, as packets may be dropped due to timeouts. We set the time-to-live (TTL) to 20 hops, and if a packet expires, its path length is counted in PSPR but not in SPR; therefore, PSPR is a metric that combines SPR and RR and can well demonstrate the performance of models.
s
\textbf{Table \ref{tab:complete}} summarises the ablation results. Our baseline model consists of GraphSAGE layers \cite{graphsage} and MLP layers, which is similar to the model Bhavanasi et al. used in their study \cite{Baseline_DRL_vs_GRL}, except that they selected GCN instead of GraphSAGE. Adding the action-space pruning mechanism augments this baseline, while replacing one linear layer with a cross-attention module provides an alternative extension. The full model combines both components, corresponding to the GRL architecture in TrajAware, as illustrated in Figure~\ref{fig:grl}.

Our experimental results show that the action-space pruning mechanism reduces PSPR and increases RR relative to the baseline under both traffic regimes. However, pruning alone remains far from optimal. Introducing cross-attention yields a substantial gain, achieving near-shortest paths and a roughly four-fold reduction in PSPR compared with pruning alone. The complete model, comprising the TrajAware GRL component, delivers the best overall reliability, achieving the best metrics in the no-congestion regime and the best PSPR and RR in the congestion mode, while also maintaining a competitive SPR. These results indicate that cross-attention is the dominant factor for accuracy, whereas pruning improves robustness and sample efficiency; together, they are necessary to obtain high delivery ratios and PSPR values close to~1 in completely unseen environments.

\subsection{Trajectory Prediction Evaluation}

We analyse our trajectory prediction module and its effect on the downstream routing task. The trajectory prediction model was also validated with the leave-one-city-out strategy to test its true generalisability. 

Recall that the trajectory prediction is recurrent; in a long-term prediction, the model treats its previous output as the next input, which inevitably leads to error accumulation. The further a vehicle is from the agent, the more outdated the observation will be. The number of steps that the trajectory prediction model must recurrently predict, namely \textit{missing steps}, is computed as:
\begin{equation}
    \textit{missing steps} = \lceil{\frac{n_{hop}}{f}}\rceil - 1
\end{equation}
Where $f$ is the broadcasting frequency and $n_{hop}$ is the hop distance between the vehicle and the agent. For example, one-hop neighbours are directly observable and thus incur no delay; two-hop neighbours require two relays, introducing a 1s delay when broadcasting once per second.

The \textit{Avg. Error} column (in meters) indicates the Euclidean distance between the predicted position and the ground truth. According to \textbf{Table \ref{tb:missing_hops}}, as the physical distance between vehicles increases, the prediction horizon becomes longer, leading to higher errors.

\begin{table}[]
\renewcommand{\arraystretch}{1.3}
    \centering
    \caption{Proportion of node pairs by geodesic distance (missing steps).}
    \begin{tabular}{|l|c|ccc|}
        \hline
        \multirow{2}{*}{\textbf{Missing Steps}} & \multirow{2}{*}{\textbf{Avg. Error}} & \multicolumn{3}{c|}{\textbf{Percentage of Occurrence}} \\ \cline{3-5}
        & & \textbf{\textit{f}=1} & \textbf{\textit{f}=2} & \textbf{\textit{f}=4} \\ \hline
        0  &  0.00  & 23.91 & 41.35 & 68.60 \\
        \hline
        1  &  23.12 & 17.45 & 27.24 & 26.13 \\
        \hline
        2  &  42.16 & 14.91 & 17.11 & 4.95 \\
        \hline
        3  &  60.05 & 12.33 & 9.02  & 0.32 \\
        \hline
        4  &  76.50 & 9.73  & 3.75  & 0.00 \\
        \hline
        5  &  92.07 & 7.38  & 1.20  & 0.00 \\
        \hline
        6  & 107.57 & 5.39  & 0.27  & 0.00 \\
        \hline
        7  & 122.86 & 3.63  & 0.05  & 0.00 \\
        \hline
        8  & 138.34 & 2.33  & 0.00  & 0.00 \\
        \hline
        9  & 153.33 & 1.42  & 0.00  & 0.00 \\
        \hline
        $\geq$10 & -- & 1.52  & 0.00  & 0.00 \\ \hline
    \end{tabular}
    \label{tb:missing_hops}
\end{table}

Since the trajectory prediction error depends on inter-vehicle distance, we examine the distribution of these distances. The three right-most columns in \textbf{Table \ref{tb:missing_hops}} report the proportion of vehicle pairs corresponding to different \textit{missing steps} values.

\begin{figure}[!t]
\centering
\includegraphics[width=3.2in]{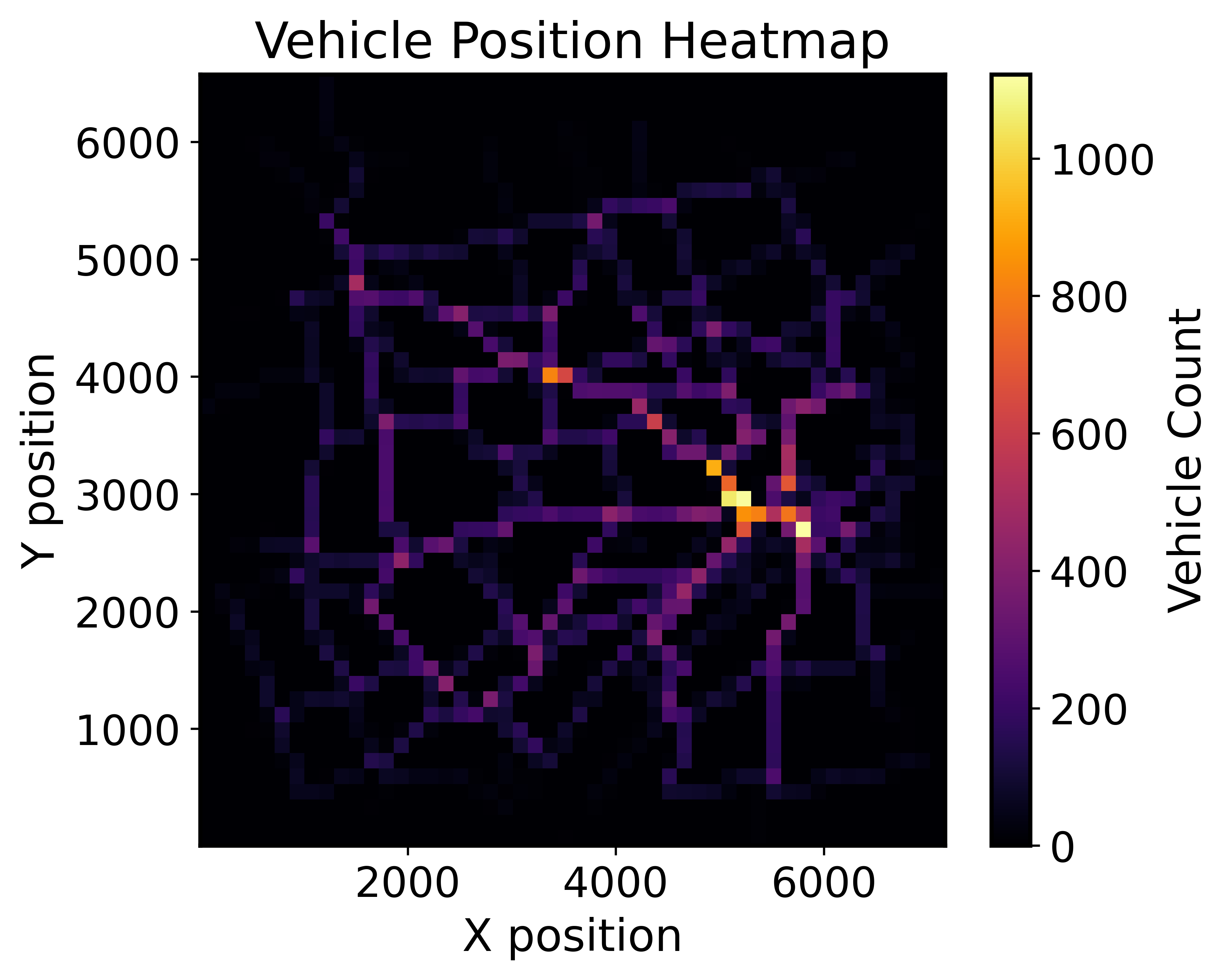}
\caption{Vehicle position heatmap illustrating the spatial distribution of traffic density. Brighter regions correspond to higher vehicle counts, indicating that traffic is predominantly concentrated in the city centre. In contrast, peripheral and suburban areas exhibit significantly lower vehicle densities.}
\label{fig:heatmap}
\end{figure}

Each non-zero missing step covers an annulus-shaped area. With a fixed width, the annulus's area enlarges as the radius increases. Although intuitively, the probability of having vehicle pairs further apart should be greater, the table showed otherwise. This is because vehicles do not distribute uniformly. As shown in the \textbf{Figure \ref{fig:heatmap}}, most vehicles cluster in the congested city centre, resulting in generally short inter-vehicle distances. In contrast, suburban areas have more sparse traffic, and longer missing steps occur less frequently. Therefore, for randomly generated routing tasks, the target is more likely to be closer to the initial sender, and thus the trajectory prediction error tends to be small.

\subsection{Incomplete Observation}
We evaluated the performance of our routing algorithm in VANET routing. We found that many related routing studies make different assumptions, making a fair and meaningful comparison difficult. Since Weil et al. \cite{Recurrent_GRL} addressed partial observation in routing problems and made their code open-source, we implemented an experimental environment based on theirs. We compared our method with four methods that they evaluated. 

All methods evaluated in the work of Weil et al. \cite{Recurrent_GRL} are based on reactive message passing, and our process is based on proactive message passing. In their setup, agents perform message passing with each routing hop. The number of message-passing steps is always no less than the shortest-path length. Our method, assuming VANET with predictable mobility, performs all message passing before routing and relies on position prediction during routing, without further message passing.

Reactive methods have an inherent drawback: at the start of routing, the agent can only observe its immediate neighbours and may need to select the next hop randomly. In contrast, proactive methods fail outright if the target’s information is unavailable before routing begins. Consequently, reactive methods generally achieve higher RR and SPR.

We focus on comparing models rather than routing strategies. To reduce the methodological gap, we allowed all reactive methods to exchange messages before routing, effectively turning them into hybrid methods. In our partially observable environment, communication occurs four times per second. Reactive methods have access to the positions of vehicles within four hops (inclusive) and acquire information about more distant neighbours through reactive broadcasting. Similarly, our proactive method also obtains positions of vehicles within four hops, but relies on previously collected data to predict the positions of vehicles beyond this range.

We follow Weil et al.’s setup. Specifically, RMP and GConv-LSTM perform one message-passing step per hop, whereas ADGN and GraphSAGE-DQN perform eight steps per hop. However, we did not adopt their replay buffer size. This is because memory consumption scales with both the network size and the maximum node degree. Although we used the same amount of RAM as in their setup (32 GB), our environment involves a larger network and a higher maximum node degree. Therefore, we reduced the replay buffer size to ensure that the experiments fit within the 32 GB memory limit.

\textbf{Table \ref{tab:incomplete}} reports five methods in the partially observed environment, RMP refers to the recurrent message passing method of Weil et al. \cite{Recurrent_GRL}. Our TrajAware model was not retrained to adapt to incomplete observation; instead, we use the model trained with complete observation and the trajectory prediction model directly. We believe that if the trajectory prediction is sufficiently accurate, the model does not need to know if the observation is incomplete.

TrajAware achieves the best results in \textbf{Table \ref{tab:incomplete}} primarily because all other models rely on the MLP layers to bridge the graph representation and the action space. Consistently, methods that omit cross-attention perform poorly in Table \ref{tab:complete}. The performance of TrajAware in Table \ref{tab:incomplete} is worse than in Table \ref{tab:complete}, particularly in terms of RR, because under partial observability, trajectory prediction may mislocalise the destination. When the predicted target position is deemed unreachable within the estimated graph structure, the agent drops the packet, which increases the drop rate and thereby elevates PSPR.

\begin{table}[!t]
\renewcommand{\arraystretch}{1.3} 
\centering
\caption{Performance comparison of different methods.}
\label{tab:incomplete}
{%
\begin{tabular}{lccc}
\hline
\multirow{2}{*}{\textbf{Method}} & \multicolumn{3}{c}{\textbf{Metrics}} \\ \cline{2-4}
 & \textbf{Avg. SPR} & \textbf{Avg. PSPR} & \textbf{RR}\\ \hline
RMP & 2.5812 & 5.0587 & 0.1362 \\
GConv-LSTM & 2.8288 & 4.7138 & 0.2175 \\
ADGN & 1.5050 & 5.1673 &  0.0959 \\
GraphSAGE-DQN & 1.4095 & 5.2070 & 0.0887 \\
TrajAware & \textbf{1.1219} & \textbf{1.2417} & \textbf{0.9307} \\ \hline
\end{tabular}%
}
\end{table}

\section{Conclusion}
In conclusion, we addressed the challenge of VANET routing under partial observability and dynamic network topology by proposing TrajAware, a novel graph neural network-based reinforcement learning framework. Our routing system combines a GraphSAGE-based GNN backbone with proactive message sharing, action-space pruning, cross-attention, and trajectory prediction to overcome key limitations of prior routing methods. Unlike earlier approaches that required retraining for new scenarios or assumed a fully visible, static network, TrajAware generalises across diverse and evolving network conditions without retraining across unseen environments. Through extensive SUMO simulations in urban environments, we demonstrated that TrajAware achieves near-optimal routing performance, maintaining average path lengths close to the shortest possible and high packet delivery ratios, significantly outperforming baseline algorithms under both complete and partial network information. An ablation analysis confirmed that both proposed components contribute to the model’s performance and robustness, enabling successful packet delivery even in completely unseen environments.

Like other proactive routing schemes, our approach inherits a fundamental limitation: when the destination is unknown, a purely proactive agent cannot select the next hop. However, TrajAware can be readily hybridised with reactive routing. For example, the RMP in \textbf{Table \ref{tab:incomplete}} can be combined with our method, allowing subsequent hops to replace estimated positions with new observations. Such a hybrid alleviates the limitation of purely proactive schemes, since only one agent along the path needs to know the destination’s position. Exploring the hybridisation of our method with other reactive approaches is therefore a promising direction for future work.

\bibliographystyle{IEEEtran}
\bibliography{bibliography}











\newpage

 




\vfill

\end{document}